\documentclass[sigconf]{aamas}

\usepackage{balance} %
\usepackage[abbreviations,british]{foreign}
\usepackage{multirow}
\usepackage{csvsimple}
\usepackage{booktabs}
\usepackage{bm}
\usepackage{caption}
\usepackage{subcaption}
\usepackage{microtype}

\setcopyright{none}
\acmConference[ALA '23]{Proc.\@ of the Adaptive and Learning Agents Workshop (ALA 2023)}
{May 29-30, 2023}{London, UK, \url{https://alaworkshop2023.github.io/}}{Cruz, Hayes, Wang, Yates (eds.)}
\copyrightyear{2023}
\acmYear{2023}
\acmDOI{}
\acmPrice{}
\acmISBN{}
\acmSubmissionID{???}

\title{Revisiting the Gumbel-Softmax in MADDPG}

\author{Callum Rhys Tilbury}
\affiliation{
  \institution{University of Edinburgh}
  \city{Edinburgh}
  \country{United Kingdom}}
\email{callum@tilbury.co.za}

\author{Filippos Christianos}
\affiliation{
  \institution{University of Edinburgh}
  \city{Edinburgh}
  \country{United Kingdom}}
\email{f.christianos@ed.ac.uk}

\author{Stefano V. Albrecht}
\affiliation{
  \institution{University of Edinburgh}
  \city{Edinburgh}
  \country{United Kingdom}}
\email{s.albrecht@ed.ac.uk}

\begin{abstract}
MADDPG is an algorithm in multi-agent reinforcement learning (MARL) that extends the popular single-agent method, DDPG, to multi-agent scenarios. Importantly, DDPG is an algorithm designed for continuous action spaces, where the gradient of the state-action value function exists. For this algorithm to work in discrete action spaces, discrete gradient estimation must be performed. For MADDPG, the Gumbel-Softmax (GS) estimator is used---a reparameterisation which relaxes a discrete distribution into a similar continuous one. This method, however, is statistically biased, and a recent MARL benchmarking paper suggests that this bias makes MADDPG perform poorly in grid-world situations, where the action space is discrete. Fortunately, many alternatives to the GS exist, boasting a wide range of properties. This paper explores several of these alternatives and integrates them into MADDPG for discrete grid-world scenarios. The corresponding impact on various performance metrics is then measured and analysed. It is found that one of the proposed estimators performs significantly better than the original GS in several tasks, achieving up to 55\% higher returns, along with faster convergence.

\end{abstract}


\newcommand{\BibTeX}{\rm B\kern-.05em{\sc i\kern-.025em b}\kern-.08em\TeX}

\begin{document}

\pagestyle{fancy}
\fancyhead{}

\maketitle

\section{Introduction}\label{sect:intro}
In recent years, interest in the field of reinforcement learning (RL) has grown markedly. Though in existence for over three decades, the discipline's recent integration with deep learning, often called deep RL, has catalysed a renewed hope for its capabilities. Such excitement is certainly warranted: deep RL algorithms have been excelling consistently on a wide range of challenges, many of which seemed unthinkable in the past. Commonly cited feats include conquering popular games, both modern and ancient~\cite{vinyalsGrandmasterLevelStarCraft2019,openaiDotaLargeScale2019,wurmanOutracingChampionGran2022,silverMasteringGameGo2016}.

An important type of problem in RL is where not only a single agent acts, but multiple agents. These agents act together, either adversarially, co-operatively, or some combination thereof. Broadly, this paradigm is termed \emph{multi-agent} RL (MARL). Algorithms developed for single-agent contexts can be applied for multiple agents, where each agent simply learns independently, \eg independent Q-learning~\citep{tan1993multi}. Though suitable for some tasks, this approach struggles to learn desired behaviours in certain complex environments, such as those with partial observability~\citep{papoudakisBenchmarkingMultiAgentDeep2021}. As an alternative, researchers have developed MARL-specific algorithms---either by extending extant single-agent approaches to multi-agent scenarios, or by developing new algorithms altogether.

One of the earliest algorithms proposed for deep MARL (that is, MARL with the integration of deep learning) was MADDPG, by~\citet{loweMultiAgentActorCriticMixed2017}. In this work, the authors extended the single-agent DDPG~\cite{lillicrapContinuousControlDeep2016} method, which is itself an extension of the DPG~\cite{silverDeterministicPolicyGradient2014} method, to multi-agent scenarios. Crucially, DPG and its descendants are designed to work only with \emph{continuous} action spaces, where each action comes from an uncountable, continuous domain; \eg the torque applied to motor. The alternative is a \emph{discrete} action space, which has countable set of possibilities; \eg choosing to go up or down. The restriction to continuous domains is because the gradient of the state-action value function, taken with respect to the action, must exist. In a discrete action context, this gradient does not exist.

Despite this restriction, it seems that the authors of MADDPG desired a unified algorithm, which could be applied to both continuous and discrete problems, while still building on the foundations of DPG. To enable MADDPG to work in discrete situations, then, a mathematical trick was applied: the \emph{Gumbel-Softmax} (GS) reparameterisation~\citep{jangCategoricalReparameterizationGumbelSoftmax,maddisonConcreteDistributionContinuous}. Essentially, this trick `relaxes' the discrete, non-differentiable action space into a somewhat equivalent, continuous space---thus allowing an approximation of the gradient to exist. Relaxing the space in this way, however, introduces statistical bias into the gradient computation. 

Recently, a benchmarking paper by~\citet{papoudakisBenchmarkingMultiAgentDeep2021} found that MADDPG achieved decent performance in certain MARL environments, but performed markedly worse in grid-world situations, where the action space is discrete. The authors suggested that this degradation of performance may be due to the bias from the GS.

Interestingly, this field of \emph{discrete gradient estimation} appears in a host of contexts outside of MARL~\citep{rolfeDiscreteVariationalAutoencoders, kusner2016gans, yan2018hierarchical}. As a result, a wealth of alternatives has been proposed for the GS, many of which focus on lowering the bias it introduces~\citep{paulusRaoBlackwellizingStraightThroughGumbelSoftmax, gappedStraightThrough, leeReparameterizationGradientNondifferentiable2018}. As of yet, though, it seems that not many of these techniques have been integrated into MARL, and certainly not into MADDPG. Accordingly, we ask:\\\emph{Can alternative discrete gradient estimation methods improve the performance of MADDPG in grid-world environments, when compared to the original Gumbel-Softmax reparameterisation?}

In this paper, we study four alternative estimators: two with simple changes to the existing GS (decreasing or annealing the relaxation temperature used), and two novel methods from the literature~\citep{paulusRaoBlackwellizingStraightThroughGumbelSoftmax, gappedStraightThrough}. We test these estimators with MADDPG in nine grid-world tasks, as a subset of those in the benchmarking paper by \citet{papoudakisBenchmarkingMultiAgentDeep2021}. We find that the Gapped Straight Through (GST)~\citep{gappedStraightThrough} estimator yields the most significant improvements, with faster convergence and up to 55\% higher returns. Finally, we look at the variance of the gradients from the GST compared to those from the GS for one of the tasks, to help us understand the performance gains observed. We release our code online\footnote{\url{https://github.com/uoe-agents/revisiting-maddpg}}.
\section{Background}
We consider a multi-agent learning problem, which we model as a partially-observable stochastic game (POSG)~\citep{shapleyStochasticGames1953, HansenPOMDP}, operating in discrete time-steps, with a set of \(N\) agents, \(\mathcal{N} = \{1,\dots,N\}\). Denote the state-space as~\(\mathcal{S}\), the joint-action space as \(\mathcal{A} = \mathcal{A}_1 \times \dots \times \mathcal{A}_N\), and the joint-observation space as \(\mathcal{O} = \mathcal{O}_1 \times \dots \times \mathcal{O}_N\). At each time-step, each agent \(i\in\mathcal{N}\) takes an action \(a_i \in \mathcal{A}_i\), and perceives a local observation, \(o_i \in \mathcal{O}_i\), which depends on the current state and the joint-action taken. We define a transition function, \(P : \mathcal{S} \times \mathcal{A} \times \mathcal{S} \mapsto [0,1]\), which describes the probability of transitioning from one state to another, given a joint action. We further define a reward function for each agent, \(\mathcal{R}_{i} : \mathcal{S}\times\mathcal{A}\times\mathcal{S}\mapsto\mathbb{R}\). Let the reward given to agent \(i\) at time-step \(t\) be denoted as \(r_i^{(t)}\). We define the \emph{return} for an agent \(i\) as its discounted cumulative reward, \(G_i = \sum_{t=0}^{T} \gamma^{t} r_i^{(t)}\), where \(T\) is the number of time-steps in an episode, and \(\gamma\in(0,1]\) is a discounting factor---controlling how much we care about future rewards relative to current rewards. The game begins in an initial state, which depends on the distribution \(\rho = \mathcal{S}\mapsto[0,1]\).

Denote each agent's policy (that is, what action it should take in a given state) as \(\pi_i\), with the set of all policies being \(\pi=\{\pi_1, \dots, \pi_N\}\). The objective in MARL, then, is to find policies such that the return of each agent \(i\), following~\(\pi_i\), is maximised with respect to the other agents' policies, \(\pi_{-i} := \{ \pi \backslash \pi_i \}\). That is, we aim to find an optimal set of policies, \(\pi\), such that
\begin{equation}\label{eqn:rl_optimal_policies}
    \forall i  :  \pi_i \in {\arg\max}_{\hat{\pi}_i}  \mathbb{E} \big[ G_i \mid \hat{\pi}_i, \pi_{-i} \big]
\end{equation}
Our focus here is on \emph{policy gradient} methods, where each agent's policy is explicitly encoded as a parametric distribution over actions given the state:  \(\pi_i(a\mid s; \theta)\). Assuming the policy is differentiable with respect to its parameters (\ie \(\frac{\partial\pi(a\mid s; \theta)}{\partial\theta}\) exists), an optimal policy can be found through gradient ascent of the expected return. To optimise this expectation, one can apply the \emph{Stochastic Policy Gradient Theorem}~\citep{suttonPolicyGradientMethods}:
\begin{equation} \label{eqn:pgt}
    \nabla_\theta \mathbb{E}_{\pi}[r(s,a)] = \mathbb{E}_{\pi}[\nabla_\theta \log \pi (a\mid s; \theta) \, Q^\pi(s,a)]
\end{equation}
where \(Q^\pi\big(s_i,a_i\big) := \mathbb{E}_\pi [G_i \mid s_i,a_i]\), the state-action value function.

Notice that the gradient, \(\nabla_\theta\), is taken with respect to the policy parameters, which exists if our policy is designed to be sufficiently smooth. This approach is simple and has been popular; \eg estimating \(Q^\pi(s,a)\) with the sampled return yields the REINFORCE algorithm~\citep{williamsSimpleStatisticalGradientfollowing}. However, it has been shown to suffer from high variance~\cite{mohamedMonteCarloGradient2020}.

Taking a different angle, \citet{silverDeterministicPolicyGradient2014} introduced the Deterministic Policy Gradient (DPG) method. Here, instead of trying to learn a stochastic policy, the policy is deterministic: \(a = \mu(s; \theta)\). The result is the \emph{Deterministic Policy Gradient Theorem}:
\begin{equation} \label{eqn:dpgt}
    \nabla_\theta \mathbb{E}_{\mu} \big[ r(s,a = \mu(s;\theta)) \big] = \mathbb{E}_{\mu}\left[ \nabla_\theta \, \mu(s;\theta) \; \nabla_a \, Q^{\,\mu}(s,a) \right]
\end{equation}
Notice the key difference now: a gradient is also taken with respect to the actions, \(\nabla_a\). Importantly, this difference means one cannot use DPG in discrete action problems, for this gradient does not exist.

DDPG~\citep{lillicrapContinuousControlDeep2016} extends the DPG method by integrating it with deep neural networks, and incorporating techniques introduced in the DQN paper~\citep{mnihPlayingAtariDeep2013}: separate `behaviour' and `target' networks with Polyak averaging, and the use of a replay buffer. MADDPG~\citep{loweMultiAgentActorCriticMixed2017} then applies DDPG to the multi-agent setting. At the heart of MADDPG is the \emph{Centralised Training, Decentralised Execution} (CTDE) paradigm. Here, each agent's state-action value function (\ie critic) is learned in a centralised manner, endowed with the joint observations and joint actions taken: \(Q_i(\bm{o}, \bm{a})\). These critic networks, parameterised by \(\phi_i\), are updated by minimising the loss, \(\mathcal{L}_c\):
\[\nabla_{\phi_i} \mathcal{L}_c(\phi_i)= \nabla_{\phi_i} \left(r_i + \gamma \bar{Q}_i \Big(\bm{o'}, \bar{\mu}_1(o_1'), \dots, \bar{\mu}_N(o_N')\Big) -Q_i(\bm{o},\bm{a})\right)^2\]
where \(\bar{\mu}_i\) denotes the target policy networks, and \(\bar{Q}_i\) the target critic networks.

Whereas the critic networks in MADDPG are centralised, notice that the policy networks are decentralised---with each agent considering only their local observation, \(\mu_i(o_i)\). These networks are updated using the sampled policy gradient:
\[-\nabla_{\theta_i}\mathcal{L}_a(\theta_i) = \nabla_{\theta_i}\,\mu_i(o_i) \; \nabla_{a_i}Q_i(\bm{o},a_1,\ldots, a_i,\ldots,a_N)\big|_{a_i= \mu_i(o_i)}\]

Notice again, as in (\ref{eqn:dpgt}), that a gradient is taken with respect to the action, \(\nabla_{a_i}Q_i\), which does not exist with discrete actions. The authors of both DPG and DDPG paid no attention to this restriction, for their methods were presented explicitly for continuous action problems. In contrast, the MADDPG algorithm was presented for both continuous and discrete cases. To enable the gradient to exist in the latter, the authors used the \emph{Gumbel-Softmax} (GS) trick~\citep{jangCategoricalReparameterizationGumbelSoftmax,maddisonConcreteDistributionContinuous} to the discrete actions taken, thus enabling an approximation of \(\nabla_{a_i}Q_i\) to exist.
\section{Discrete Gradient Estimation}
In this work, we consider four alternatives to the original GS trick: two of which apply simple changes to the extant method, and two of which are novel methods drawn from the literature. We now provide brief explanations of these techniques, starting with the original GS method.

We consider a situation of a parametric discrete distribution, \(p(a; \zeta)\), specified by an unconstrained vector of parameters, \(\zeta \in \mathbb{R}^N\). In this context, these parameters represent the outputs of a policy network, where actions are sampled as \(a\sim p(a;\zeta)\). We seek an estimation of \(\nabla_a\, p(a;\zeta)\), but since \(a\) is discrete, we must `relax' the distribution for this gradient to exist.

\subsection{Baseline (STGS-1)}
The GS method was introduced concurrently by~\citet{jangCategoricalReparameterizationGumbelSoftmax} and~\citet{maddisonConcreteDistributionContinuous}, as a differentiable approximation of the \(\arg\max\) function. As its name suggests, a tempered \emph{softmax} is used, with a temperature parameter, \(\tau>0\): \(\textrm{softmax}_\tau (x) := \textrm{softmax} (\frac{x}{\tau})\). In the limit of \(\tau \to 0\), this operation is equivalent to the \(\arg\max\), and thus the GS approaches the original distribution. Conversely, as \(\tau \to \infty\), the GS approaches a uniform distribution, where each category is equally-likely. The temperature thus controls the degree of relaxation.

Using \(\xi_{GS}(\cdot)\) to notate the GS, the relaxed distribution is:
\begin{align}
    \xi_{\textrm{GS}}(p(a_i;\zeta)) &= \textrm{softmax}_\tau (\zeta_i + g_i) \quad,\quad g_i \sim G(0,1)%
\end{align}
where \(g_i\) is noise sampled from the Gumbel distribution~\citep{pml2Book}.

By relaxing the distribution in this way, it becomes differentiable, meaning we can incorporate it into a gradient-based optimisation procedure. There is a downside, however: in relaxing, we introduce statistical bias~\citep{paulusRaoBlackwellizingStraightThroughGumbelSoftmax}. To understand this bias intuitively, consider again the limit \(\tau\to\infty\), where the distribution becomes uniform. In such a case, we have removed all parametric information, \(\zeta\), about our problem---each category simply has a probability of \(1/N\). Hence, as we relax, we also steer further away from the original distribution. Herein lies a trade-off: turning the temperature too low means having extreme gradients (or non-existent gradients when \(\tau=0\)), but turning the temperature too high means introducing a large bias.

Though such a bias is inevitable when relaxing the distribution, there is an easy improvement to the vanilla GS. By naïvely applying the relaxation, we introduce bias in both the forward pass (when we sample from the distribution) and in the backward pass (when we calculate the gradients, \eg for updating our neural network). However, it is only the latter that requires differentiability. Hence, building on the so-called \emph{Straight-Through} estimator proposed by~\citet{bengioGradients}, \citet{jangCategoricalReparameterizationGumbelSoftmax} also introduce the STGS estimator---where, in the backward pass, the GS relaxation is applied, but in the forward pass, the original \(\arg\max\) operation is used. MADDPG uses this variant, and all further discussions focus on it.

In both the original MADDPG paper~\citep{loweMultiAgentActorCriticMixed2017} and the benchmarking paper by~\citet{papoudakisBenchmarkingMultiAgentDeep2021}, it seems that the authors simply use a temperature of \(1.0\) for the STGS relaxation\footnote{Nothing is explicitly stated about the temperature used in these papers; we are making such conclusions by looking at their code implementations: \href{https://github.com/openai/maddpg/blob/3ceefa0ada3ff31d633dd0bde8ff95213ce99be3/maddpg/common/distributions.py\#L205}{Link to snippet} from~\citet{loweMultiAgentActorCriticMixed2017}; \href{https://github.com/uoe-agents/epymarl/blob/96db475082b7227f295b927927654b2dd91d80d4/src/learners/maddpg_learner.py\#L95}{Link to snippet} from~\citet{papoudakisBenchmarkingMultiAgentDeep2021}.}. Thus, we use this configuration as our baseline, denoted as \textbf{STGS-1}.

\subsection{Lower Temperature Gumbel Softmax (STGS-T)}
Recall that \citet{papoudakisBenchmarkingMultiAgentDeep2021} suggest it is the bias that is problematic in the STGS, which is positively correlated to the temperature of relaxation---lowering the temperature lowers the bias. Accordingly, our first alternative estimator is the simplest: the STGS estimator with a temperature of \(\tau < 1.0\), where \(\tau\) is a tunable hyperparameter, denoted as \textbf{STGS-T}.

\subsection{Temperature-Annealed Gumbel Softmax (TAGS)}
The \emph{exploitation-exploration dilemma}~\citep{suttonReinforcementLearningIntroduction2018}---which describes the trade-off between taking actions that yield known, good rewards (exploiting), and taking actions which may or may not yield better rewards (exploring)---is often discussed in RL. In the continuous-action formulation of MADDPG~\citep[Appx: Alg. 1]{loweMultiAgentActorCriticMixed2017}, exploration is achieved via the addition of noise to the policy output: \(a_i = \mu_i(o_i) + \eta_i\), where \(\eta\) is drawn from some random process (originally discussed in DDPG~\citep{lillicrapContinuousControlDeep2016}). However, in discrete cases, the STGS itself provides some degree of exploration, since relaxing the distribution places some probability mass onto other actions. As a result, the amount of exploration is controlled by the temperature parameter: more relaxation implies more exploration. Notice, then, the coupling between the exploration achieved and the bias introduced.

Since exploration is usually desirable in the beginning of a training procedure, we propose setting the temperature to be high early-on, and then annealing it to be lower over time. This strategy allows agents to explore, while still reducing the bias in later stages of training. \citet{huijbenReviewGumbelmaxTrick2022} highlight temperature-annealing as a strategy incorporated by several authors in various experiments with the STGS. Specifically, they mention using an exponentially-decaying annealing scheme, which we adopt here. We define this estimator as the Temperature-Annealed Gumbel Softmax (\textbf{TAGS}).

\subsection{Gumbel-Rao Monte Carlo (GRMC\(K\))}
The next estimator is drawn from the literature, entitled the Gumbel-Rao Monte Carlo (GRMC), by~\citet{paulusRaoBlackwellizingStraightThroughGumbelSoftmax}. Here, the authors seek a way to lower the STGS's variance. Notating the gradient of the original STGS estimator as \(\nabla_{\textrm{STGS}} = \nabla_a \xi_{\textrm{STGS}}\), we have:
\begin{equation}
    \nabla_{\textrm{STGS}} := \frac{d\textrm{softmax}_\tau (\zeta + g)}{da}
\end{equation}
With this notation, the authors propose the Gumbel-Rao estimator:
\begin{equation}
    \nabla_{\textrm{GR}} := \mathbb{E}\left[\frac{d\textrm{softmax}_\tau (\zeta + g)}{da} \;\Big|\; a \right]
\end{equation}

That is, \(\nabla_{\textrm{GR}}=\mathbb{E}[\nabla_{\textrm{STGS}}\mid a]\). This estimator is a Rao-Blackwell~\citep{blackwell1947conditional} version of the original STGS estimator. It can be shown that it thus enjoys the same mean as the STGS, but with lower (or at most, the same) variance:
\begin{equation}
    \mathbb{E}\left[||\nabla_{\textrm{GR}} - \nabla_{\zeta}||^2\right] \le \mathbb{E}\left[||\nabla_{\textrm{STGS}} - \nabla_{\zeta}||^2\right]
\end{equation}
where \(\nabla_{\zeta}\) is the true gradient. For rigorous mathematical details about the estimator's impact on variance, the reader is encouraged to see the full paper~\citep{paulusRaoBlackwellizingStraightThroughGumbelSoftmax}. Recall that \citet{papoudakisBenchmarkingMultiAgentDeep2021} consider the bias of the estimator to be the problem for MADDPG, not the variance. Though guarantees are only made about the latter, the authors of the GRMC argue that with a lower variance, one can safely train the estimator at lower temperatures---\ie with a lower bias. Empirically, they show this idea to be true.

Though theoretically appealing, there is still the challenge of actually computing \(\mathbb{E}[d\textrm{softmax}_\tau (\zeta + g) / d a \mid a]\)---indeed, a closed-form expression is shown to be difficult. Therefore, the authors provide a Monte Carlo estimate, with \(K\) samples, which they term the \textbf{GRMC\(K\)} estimator. They first show a distributional equivalence:
\begin{equation}
    (\zeta_j + g_j \mid a) \stackrel{d}{=}
    \begin{cases}
        -\log (E_j) + \log Z(\zeta) &\quad \textrm{if j = i}\\
        -\log\left(\frac{E_j}{\exp(\zeta_j)} + \frac{E_i}{Z(\zeta)}\right) &\quad \textrm{otherwise}
    \end{cases}
\end{equation}
where \(a\) is a one-hot sample with a \(1\) at index \(i\), \(E_j\) are i.i.d. samples from the exponential distribution, and \(Z(\zeta) = \sum_j \exp(\zeta_j)\).

They then define the GRMC\(K\) estimator as:
\begin{equation}
    \nabla_{\textrm{GRMC}K} :=  \frac{1}{K} \sum_k^K \frac{d\textrm{softmax}_\tau (\zeta + g_k)}{da} \quad,\quad g_k \sim (\zeta + g \mid a)
\end{equation}

In other words, we first sample \(a\sim p(a;\zeta)\), and then average over \(K\) Gumbel noise samples conditioned on \(a\).

\subsection{Gapped Straight-Through (GST)}
The final estimator considered is the most recently introduced: the Gapped Straight Through (GST), by~\citet{gappedStraightThrough}. Here, the authors find that the Gumbel randomness used in the STGS and GRMC\(K\) can be replaced with two deterministic perturbations, resulting in an estimator with lower variance.

As in the GRMC estimator, we first draw \(a\sim p(a;\zeta)\)---a one-hot representation of the selected action---for the straight-through sample. In GRMC, we would then perturb each of the logits, \(\zeta\), with Gumbel noise conditioned on \(a\); now, we perturb with two functions, \(m_1(\zeta,a)\) and \(m_2(\zeta,a)\). Detailed justifications of these functions are given in the paper itself~\citep{gappedStraightThrough}.

Firstly, we desire \emph{consistency} in the estimator: we want the sample conditioned on~\(a\) to have the same largest logit as the input distribution; \ie \(\max_j \zeta_j\). To this end, the first perturbation, \(m_1\), pushes the sample to the correct realisation, if necessary:
\begin{equation}
    m_1(\zeta, a) = (\max_j \zeta_j - \langle \zeta, a \rangle ) \cdot a
\end{equation}
where \(\langle \cdot, \cdot \rangle\) indicates the inner product. Consider how this works: if \(a\) has already selected the largest logit, then \(\langle \zeta, a \rangle = \max_j \zeta_j\), and \(m_1 = 0\). If not, then \(m_1 \ne 0\), and the sample is moved in the direction of the largest logit.

If non-zero, the first perturbation makes the largest logit the same as the \(a\)-selected logit. However, we also want a \emph{strict gap} between these values---that is, we want the unselected logits to be smaller. Accordingly, we define \(m_2\) to create a gap of \(\kappa\) between them:
\begin{equation}
    m_2(\zeta, a) = - (\kappa + \zeta - \max_j \zeta_j)_+ \odot (1 - a)
\end{equation}
where \((x)_+ := \max(0,x)\), \(\odot\) indicates the Hadamard product. The value of \(\kappa\) can usually be set to~\(1.0\)~\citep{gappedStraightThrough}. Here, the term \((1-a)\) takes all the unselected logits in the one-hot representation, and moves their parameter values away from the selected logit, with a gap of at least~\(\kappa\).

With these perturbation functions defined, the GST estimator is then:
\begin{equation}
    \xi_{\textrm{GST}}(p(a; \zeta)) = \textrm{softmax}_\tau \left(\zeta + m_1(\zeta, a) + m_2(\zeta, a)\right)
\end{equation}

\section{Experimental Methods}
\subsection{Environments}
To test the performance of the proposed gradient estimators compared to the original STGS estimator, we train on two grid-world environments, across a total of nine tasks (\ie configurations). For simplicity, we choose to focus solely on co-operative contexts, where agents are working together to maximise their cumulative reward---we feel this approach is sufficient to answer our research question. We use a sensible subset of the choices made by~\citet{papoudakisBenchmarkingMultiAgentDeep2021} in their benchmarking paper: seven tasks in Level-Based Foraging (\textbf{LBF})\footnote{Code: \href{https://github.com/semitable/lb-foraging}{https://github.com/semitable/lb-foraging}}~\citep{gameTheoreticAlbrecht, christianosSharedExperienceActorCritic}, and two in Multi-Robot Warehouse (\textbf{RWARE})\footnote{Code: \href{https://github.com/semitable/robotic-warehouse}{https://github.com/semitable/robotic-warehouse}}~\citep{albrecht2016exploiting, christianosSharedExperienceActorCritic}.

\subsection{Evaluation Metrics}
To understand the success (or failure) of the new gradient estimation techniques compared to the original STGS approach, metrics for evaluation are now defined.

\textbf{Returns: Maximum \& Average.}
Recall that we define our MARL goal, in~(\ref{eqn:rl_optimal_policies}), as trying to find an optimal set of policies, such that each agent maximises their expected return with respect to the other agents' policies. Since we focus on co-operative situations for this paper, we simply consider the sum of the achieved returns from all agents. Importantly, we are not concerned with the returns achieved here relative to those achieved in, \eg, the MARL benchmarking paper by~\citet{papoudakisBenchmarkingMultiAgentDeep2021}. Instead, for cogent and consistent analysis, we focus solely on the relative performance of the various estimators against each other. 

For this evaluation metric, we run the MADDPG algorithm in each task, with each of the proposed gradient estimators. We train the algorithm for a fixed number of time-steps, updating the networks with a defined period. Throughout training, we evaluate the achieved returns \(100\) times every \(50\,000\) time-steps. Each training iteration is done over five random seeds and a 95\% confidence interval is calculated over the results.

Under this heading, we consider two distinct aspects of the achieved returns, following the lead of~\citet{papoudakisBenchmarkingMultiAgentDeep2021}. Firstly, we consider the \emph{maximum} return: the evaluation time-step at which the return, averaged over the five seeds, is highest---indicating the peak performance of the algorithm when using a given estimator. Secondly, we consider the \emph{average} return: the mean of the evaluation returns over all time-steps and seeds, for a given estimator in a given task---a proxy for understanding not just the magnitude of the returns, but how quickly the training converges.

\textbf{Compute Time.}
Though this paper revolves around---and is motivated by---the MADDPG algorithm, notice that the gradient estimators can also be compared in isolation. That is, when comparing the computational burden of the various estimation procedures, we need not integrate them into the broader MADDPG problem. Instead, we can take a closer look solely at each estimator's performance, unhindered by potential bottlenecks elsewhere.

Accordingly, we define here a simple, toy problem for the estimators. We define a set of input logits, \(\zeta\), of various dimensionalities, and measure the time it takes for each estimator to calculate the corresponding relaxations. Because STGS-1, STGS-T, and TAGS all have the same underlying mechanics, we consider these under the single umbrella of the STGS. For the GRMC\(K\), we consider three values of \(K\): 1, 10, and 50. For each dimensionality, the estimation procedure is repeated \(10\;000\) times, over five different logit instances. These results are reported over a 95\% confidence interval.

\textbf{Gradient Variance.}
Suppose one of the alternative gradient estimation techniques performs significantly better or worse than the original STGS, based on the returns achieved. The natural follow-up question is: why? As an initial step to answering this question, we choose one of the tasks where there is a notable difference in performance between two estimators: between the baseline STGS-1 method and one which performs much better (or worse). We then retrain the MADDPG algorithm in this task, using each of the two estimators, now logging the variance of the computed gradients across each training mini-batch, over the course of training. 

We hypothesise that \emph{uninformative} gradients, \ie those due to a poor discrete gradient estimator, will yield a mini-batch with low variance, since there are no elements in particular which `stand out'. In contrast, we believe that \emph{informative} gradients will have higher variance across the mini-batch, for the opposite reason. We hope for this metric to stimulate future discussion into why we might observe a difference in estimator performance.

\subsection{Training Details}
Hyperparameter tuning is an important, though time-consuming, component of training RL algorithms. For simplicity, then, the optimal hyperparameters for the core MADDPG algorithm suggested by~\citet{papoudakisBenchmarkingMultiAgentDeep2021} are adopted here, mostly without any changes. Our hyperparameter search is thus limited to be over the novel gradient estimation techniques and their associated parameters. Bayesian optimisation~\citep{garnett_bayesoptbook_2022} is performed for this search, and we use search-range suggestions from the literature, when available~\citep{huijbenReviewGumbelmaxTrick2022,paulusRaoBlackwellizingStraightThroughGumbelSoftmax}. Each parameter is optimised for one task in a particular environment, and then used for all other tasks in that environment. The resulting hyperparameters can be found online\footnote{\url{https://github.com/uoe-agents/revisiting-maddpg}}, along with the code used for all experiments.

\section{Experimental Results \& Discussion\label{chpt:experiments}}

\subsection{Returns: Maximum \& Average}
Table~\ref{tab:core-results} shows the maximum and average returns achieved using each gradient estimation technique, across each of the nine tasks. Discussion of these results follows, per environment, with plots shown when relevant.

\newcommand{\FormatReturnResult}[4]{
    \begin{tabular}{@{}c@{}}
        \ensuremath{{#1}\pm{#2}}\\[-11pt]
        (\ensuremath{{#3}\pm{#4}})
    \end{tabular}
}
\newcommand{\FormatTaskDisplay}[2]{
    \begin{tabular}{@{}c@{}}
        #1\\[-11pt] {\footnotesize[#2]}
    \end{tabular}
}
\begin{table}[htbp]
    \caption{Maximum returns (Average returns) shown across all tasks and all algorithms, presented with a 95\% confidence interval over 5 seeds. \textbf{Bold} indicates the best performing metric for a situation. An asterisk (\(*\)) indicates that a given metric is \emph{not} significantly different from the best performing metric in that situation, based on a heteroscedastic, two-sided t-test with 5\% significance. Under each task name is the number of time-steps used for training.}
    \label{tab:core-results}
    \centering
    \renewcommand{\arraystretch}{1.8}
    \setlength{\tabcolsep}{5pt}
    \resizebox{\linewidth}{!}{%
        \begin{filecontents*}{data/returns.csv}
1,theclass,theTrainingSteps,theenvironment,shouldwerule,bestMAX,bestAVG,theSTGSmax,theSTGSmaxErr,theSTGSavg,theSTGSavgErr,theSTGSTmax,theSTGSTmaxErr,theSTGSTavg,theSTGSTavgErr,theTAGSmax,theTAGSmaxErr,theTAGSavg,theTAGSavgErr,theGRMCKmax,theGRMCKmaxErr,theGRMCKavg,theGRMCKavgErr,theGSTmax,theGSTmaxErr,theGSTavg,theGSTavgErr
4,LBF,5M,8x8-2p-2f-c,x,,,\bm{1.00},\bm{0.00},0.88,0.05,\bm{1.00},\bm{0.01},\bm{0.91},\bm{0.04},\bm{1.00},\bm{0.01},0.87,0.05,\bm{1.00},\bm{0.00},0.88,0.05,\bm{1.00},\bm{0.00},0.89,0.04
5,,5M,8x8-2p-2f-2s-c,,,,0.79,0.07^*,0.65,0.04,\bm{0.83},\bm{0.03},0.66,0.04,0.78,0.03^*,0.62,0.04,0.81,0.05^*,0.67,0.04,0.81,0.02^*,\bm{0.68},\bm{0.03}
6,,6M,10x10-3p-3f,,,,0.75,0.03,0.58,0.04,0.75,0.03,0.59,0.03,0.74,0.06,0.51,0.04,0.71,0.07,0.57,0.03,\bm{0.79},\bm{0.04},\bm{0.66},\bm{0.03}
7,,6M,10x10-3p-3f-2s,,,,0.55,0.05^*,0.48,0.01,\bm{0.58},\bm{0.06},0.48,0.01,0.54,0.03,0.46,0.01,0.56,0.03^*,0.49,0.01,0.56,0.05^*,\bm{0.50},\bm{0.01}
8,,7.5M,15x15-3p-5f,,,,0.24,0.02,0.12,0.01,0.28,0.06,0.15,0.01,0.20,0.03,0.08,0.01,0.26,0.05,0.16,0.01,\bm{0.31},\bm{0.04},\bm{0.20},\bm{0.01}
9,,7.5M,15x15-4p-3f,,,,0.79,0.03,0.54,0.04,0.79,0.06,0.58,0.04,0.77,0.06,0.45,0.04,0.79,0.04,0.58,0.04,\bm{0.83},\bm{0.04},\bm{0.67},\bm{0.03}
10,,7.5M,15x15-4p-5f,,,,0.33,0.06,0.13,0.02,0.46,0.12,0.22,0.02,0.24,0.05,0.10,0.01,0.43,0.06,0.21,0.02,\bm{0.48},\bm{0.06},\bm{0.30},\bm{0.02}
11,RWARE,7.5M,tiny 2ag,x,,,1.37,0.22^*,0.55,0.07,1.37,0.49^*,0.64,0.08,\bm{1.50},\bm{0.46},0.60,0.08,1.37,0.40^*,\bm{0.65},\bm{0.07},1.40,0.58^*,\bm{0.65},\bm{0.07}
12,,7.5M,tiny 4ag,,,,2.68,0.49,0.84,0.12,3.18,0.60,1.09,0.15,2.23,0.50,0.78,0.11,3.17,0.85,1.15,0.15,\bm{4.16},\bm{0.97},\bm{1.82},\bm{0.21}
\end{filecontents*}
\csvreader[
            head to column names,
            tabular=c|c|ccccc,
            table head= \toprule & Tasks & \textbf{STGS-1} & \textbf{STGS-T} & \textbf{TAGS} & \textbf{GRMC\(K\)} & \textbf{GST},
            late after line=,
            before first line=,
            before line=\ifthenelse{\equal{\shouldwerule}{}}{\\}{\\\hline\hline},
            table foot = \\\bottomrule,
        ]{data/returns.csv}{}{%
            \ifthenelse{\equal{\theclass}{}}{}{\multirow{1}{*}{\rotatebox[origin=c]{90}{\emph{\theclass}}}} &
            \FormatTaskDisplay{\theenvironment}{\theTrainingSteps} &
            \FormatReturnResult{\theSTGSmax}{\theSTGSmaxErr}{\theSTGSavg}{\theSTGSavgErr} &
            \FormatReturnResult{\theSTGSTmax}{\theSTGSTmaxErr}{\theSTGSTavg}{\theSTGSTavgErr} &
            \FormatReturnResult{\theTAGSmax}{\theTAGSmaxErr}{\theTAGSavg}{\theTAGSavgErr} &
            \FormatReturnResult{\theGRMCKmax}{\theGRMCKmaxErr}{\theGRMCKavg}{\theGRMCKavgErr} &
            \FormatReturnResult{\theGSTmax}{\theGSTmaxErr}{\theGSTavg}{\theGSTavgErr}
        }
    }
\end{table}

\textbf{Level-Based Foraging.}
We consider now the seven LBF tasks. Firstly, we look at two tasks with two agents over an \(8\times8\) grid: one with full-observability (8x8-2p-2f), and one with partial-observability (8x8-2p-2f-2s). Notice in these results, in Table~\ref{tab:core-results}, that performance differences across the estimation techniques is statistically insignificant. In 8x8-2p-2f, we see that STGS-T trains marginally faster than the other approaches, and in 8x8-2p-2f-2s, we see that TAGS trains marginally slower than the other approaches---both based on the average returns observed. Nonetheless, each algorithm arrives at a similar maximum return. Due to the insignificance of this result, the training curves are uninteresting, and are not plotted here.

\begin{figure}[htb]
    \centering
    \captionsetup{type=figure}
    \begin{subfigure}[t]{\linewidth}
        \centering
        \includegraphics[width=\linewidth]{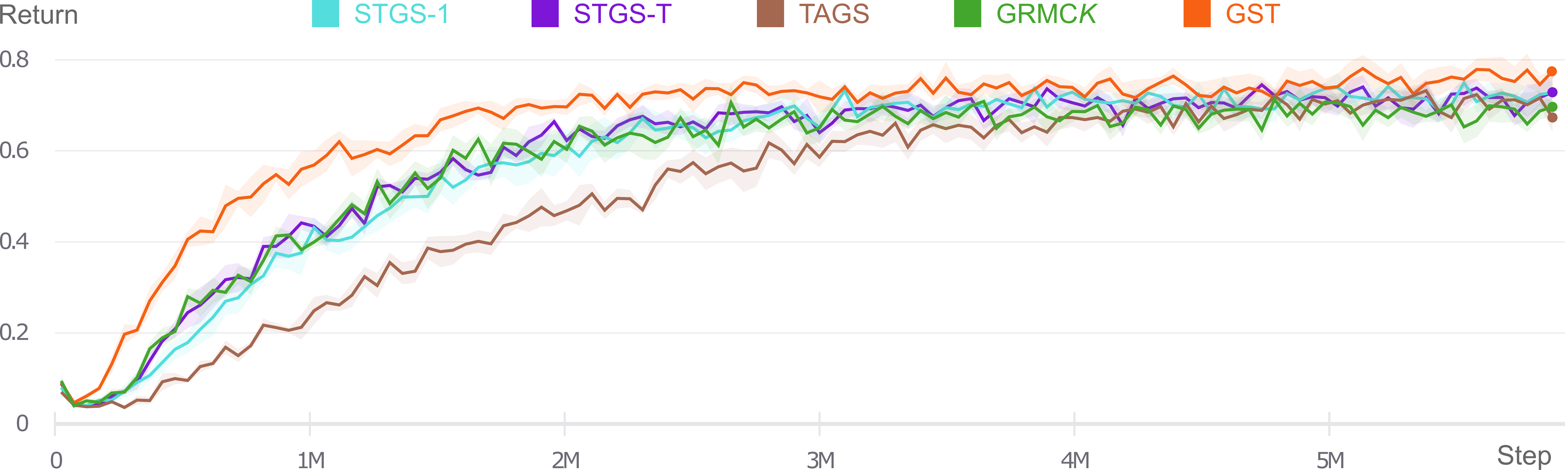}
        \caption{\texttt{lbf-10x10-3p-3f}}
        \label{fig:returns:lbf-10x10-3p-3f}
    \end{subfigure}%
    \vspace{0.3cm}
    \begin{subfigure}[t]{\linewidth}
        \centering
        \includegraphics[width=\linewidth]{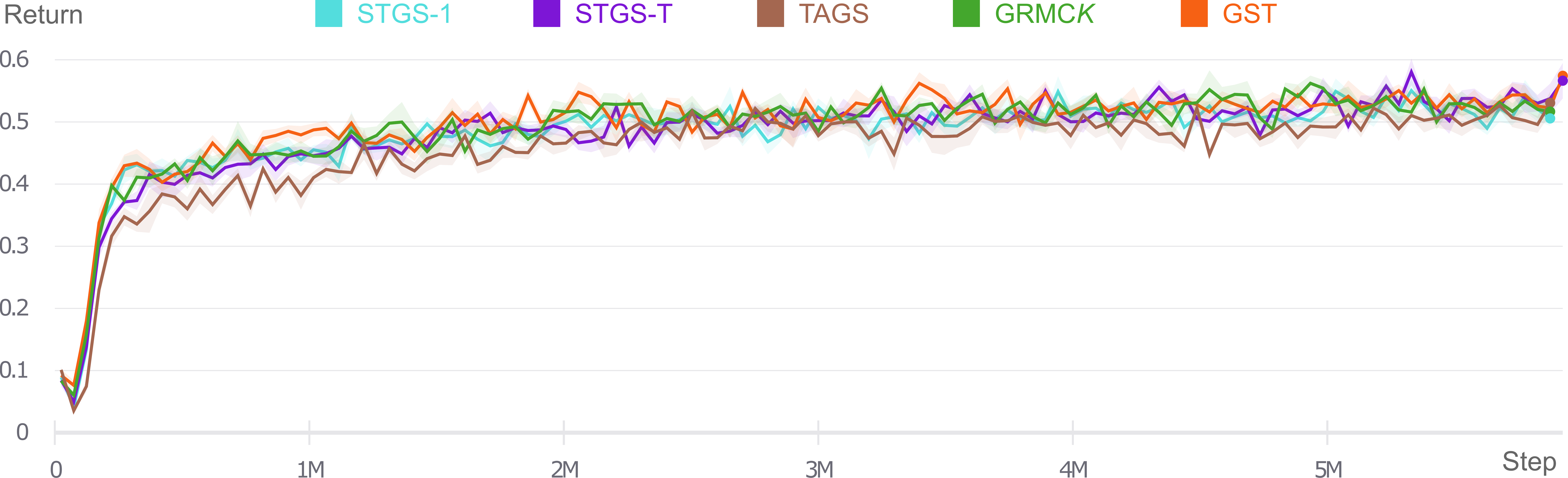}
        \caption{\texttt{lbf-10x10-3p-3f-2s}}
        \label{fig:returns:lbf-10x10-3p-3f-2s}
    \end{subfigure}%
    \caption{Evaluation returns for two LBF tasks \((10\times10)\) over the training period, where the shaded region indicates the standard error as calculated over 5 seeds.}
    \label{fig:returns:lbf-10x10}
\end{figure}

We next look at two tasks with three agents over an \(10\times10\) grid, with similar situations as before: one with full-observability (10x10-3p-3f), and one with partial-observability (10x10-3p-3f-2s). Plots of the evaluation returns over the duration of training are given in Figure~\ref{fig:returns:lbf-10x10}.

In the first task, seen in Figure~\ref{fig:returns:lbf-10x10-3p-3f}, we see an improvement with a novel gradient estimation technique: the GST achieves the highest maximum and average returns for the task, beating the baseline STGS-1 method with statistical significance. It is clear in the figure how training with the GST converges faster than with the other methods. STGS-T and GRMC\(K\) perform similarly to the baseline. TAGS, however, performs much worse in average returns---\ie it converges slower for the task---though it eventually achieves similar maximum returns.

In the task with partial observability, seen in Figure~\ref{fig:returns:lbf-10x10-3p-3f-2s}, we are less successful---the alternative techniques achieve statistically similar returns to the baseline, across both maximum and average metrics. The exception is TAGS, which again performs \emph{worse} than the baseline, though not markedly so.

\begin{figure}[htbp]
    \centering
    \captionsetup{type=figure}
    \begin{subfigure}[t]{\linewidth}
        \centering
        \includegraphics[width=\linewidth]{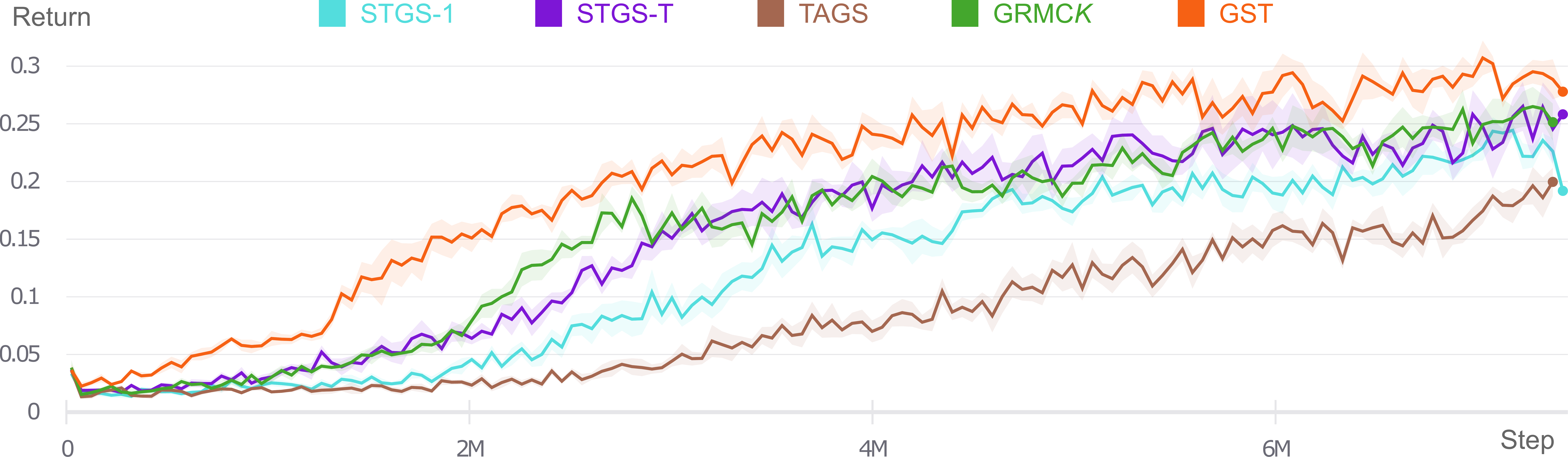}
        \caption{\texttt{lbf-15x15-3p-5f}}
        \label{fig:returns:lbf-15x15-3p-5f}
    \end{subfigure}%
    \vspace{0.3cm}
    \begin{subfigure}[t]{\linewidth}
        \centering
        \includegraphics[width=\linewidth]{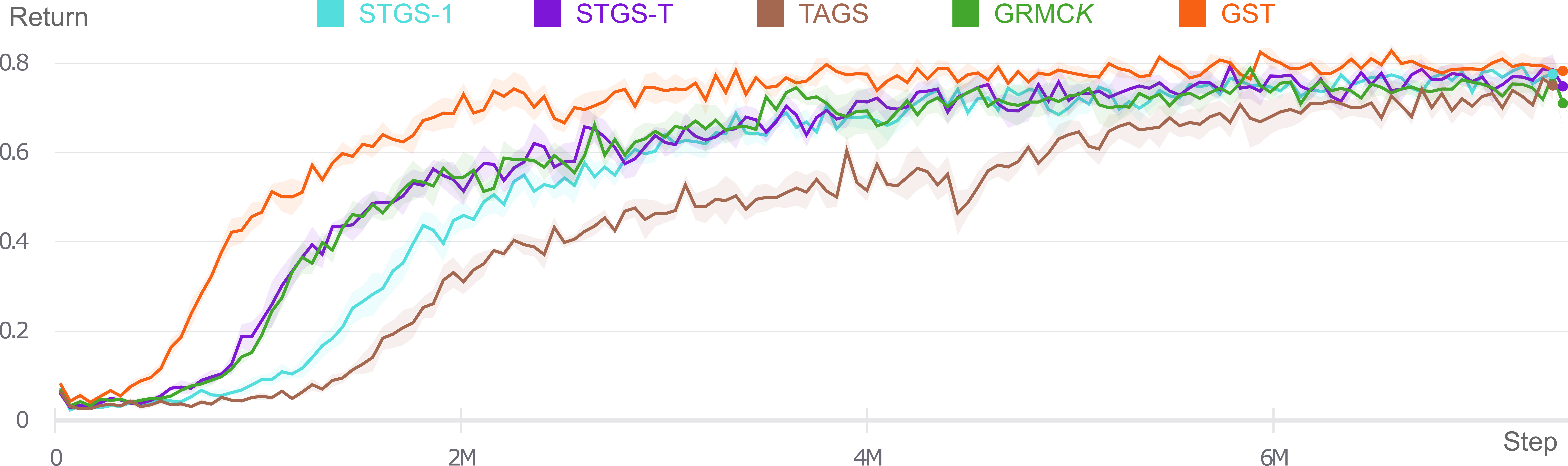}
        \caption{\texttt{lbf-15x15-4p-3f}}
        \label{fig:returns:lbf-15x15-4p-3f}
    \end{subfigure}%
    \vspace{0.3cm}
    \begin{subfigure}[t]{\linewidth}
        \centering
        \includegraphics[width=\linewidth]{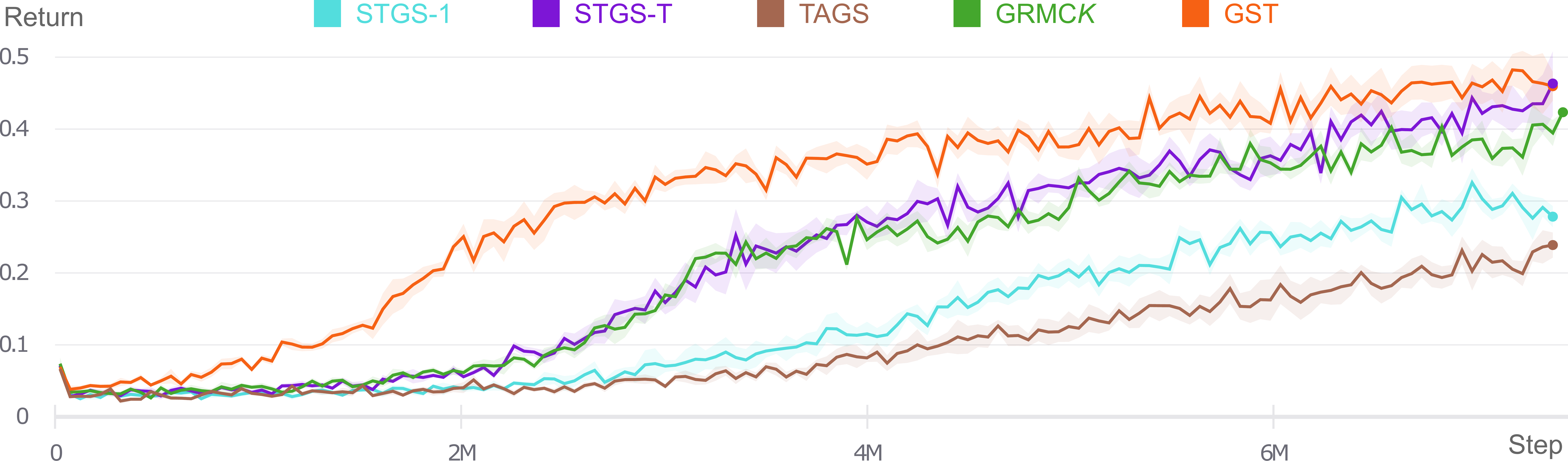}
        \caption{\texttt{lbf-15x15-4p-5f}}
        \label{fig:returns:lbf-15x15-4p-5f}
    \end{subfigure}%
    \caption{Evaluation returns for three LBF tasks \((15\times15)\) over the training period, where the shaded region indicates the standard error as calculated over 5 seeds.}
    \label{fig:returns:lbf-15x15}
\end{figure}

We consider now the remaining three tasks in LBF, with a fully-observable, \(15\times 15\) grid, with three or four agents: 15x15-3p-5f, 15x15-4p-3f, and 15x15-4p-5f. We note that MADDPG performed particularly poorly in these larger, more-complex LBF situations, according to the benchmarking paper by~\citet{papoudakisBenchmarkingMultiAgentDeep2021}. The training curves for each of these tasks is given in Figure~\ref{fig:returns:lbf-15x15}.

We notice here significant improvements over the baseline. Yet again, TAGS markedly underperforms, both in maximum and average returns; but the other estimators perform well. STGS-T and GRMC\(K\) beat the baseline in average returns for 15x15-3p-5f and 15x15-4p-3f, and in both average and maximum returns for 15x15-4p-5f. GST is superior throughout: across all three tasks, it yields significantly higher returns and converges faster than the baseline (and the other techniques). Indeed, these improvements are clearly noticeable in the plots provided.

\begin{figure}[htb]
    \centering
    \captionsetup{type=figure}
    \begin{subfigure}[t]{\linewidth}
        \centering
        \includegraphics[width=\linewidth]{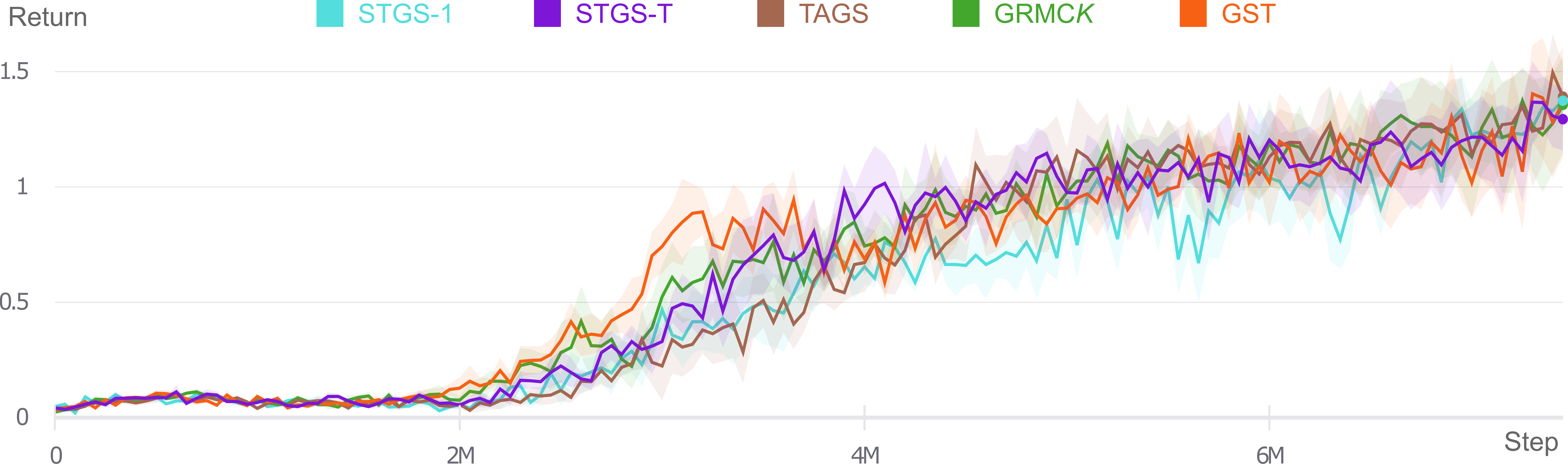}
        \caption{\texttt{rware-tiny-2ag}}
        \label{fig:returns:rware-tiny-2ag}
    \end{subfigure}%
    \vspace{0.3cm}
    \begin{subfigure}[t]{\linewidth}
        \centering
        \includegraphics[width=\linewidth]{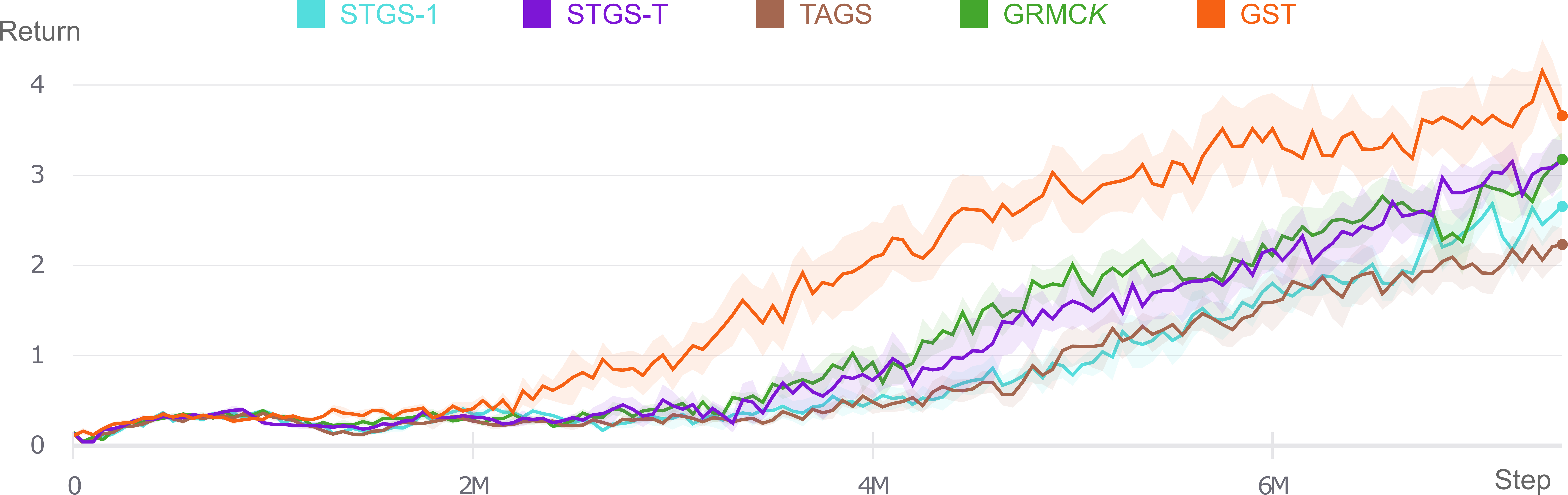}
        \caption{\texttt{rware-tiny-4ag}}
        \label{fig:returns:rware-tiny-4ag}
    \end{subfigure}%
    \caption{Evaluation returns for two RWARE tasks (tiny grid) over the training period, where the shaded region indicates the standard error as calculated over 5 seeds.}
    \label{fig:returns:rware}
\end{figure}

\textbf{Multi-Robot Warehouse.}
Next, we consider the RWARE environment for two tasks over a \(10\times11\) grid: one with two agents (tiny-2ag) and one with four agents (tiny-4ag). Figure~\ref{fig:returns:rware} shows the returns for these two environments, over the training period.

In tiny-2ag, we see insignificant differences across the estimation techniques, with each achieving similar maximum returns. The alternative techniques do converge slightly faster, particularly GRMC\(K\) and GST, with marginally higher average returns, but not by much.

In tiny-4ag, we see the most significant improvements yet. Barring TAGS, which somewhat underperforms, we notice substantial improvements from the other proposed estimators, for both average and maximum returns. GST triumphs once more, achieving 55\% higher maximum returns over the baseline, and over double the average returns. This result is again clear in the plot, in Figure~\ref{fig:returns:rware-tiny-4ag}.

\textbf{Discussion.} This section presented the results from training MADDPG with each of the proposed gradient estimation techniques, across nine tasks from two grid-world environments. In simpler tasks, the alternative techniques do not make a significant difference to the returns achieved. We suspect this outcome is because informative gradients are not as crucial in simple tasks. That is, the gradient estimation is not a problematic aspect of the training, and limitations arise elsewhere in the mechanics of MADDPG.

Interestingly, in some of the more challenging tasks, particularly in LBF with a grid-size of \(15\times15\), and tiny-4ag in RWARE, we see significant improvements. We note that simply lowering the temperature (and hence, the gradient estimator's bias), as in STGS-T, can improve the results somewhat---supporting the hypothesis that the bias introduced by the STGS is a problem for MADDPG. The Rao-Blackwellisation procedure of GRMC\(K\) also sees better returns and faster convergence. Much better than these alternatives, though, is the GST. With this estimator, we consistently see marked improvements across the two return metrics, and these are statistically significant.

Though a lower temperature seems to yield better returns, our results suggest that annealing the temperature, as in TAGS, performs poorly. This result may be due to the coupling of exploration and exploitation, as highlighted earlier, but more investigation is required. Alternatively, it may simply be the hyperparameters chosen---the annealing start and end points were taken from the advice of~\citet{huijbenReviewGumbelmaxTrick2022}. It is conceivable that using lower values here may yield better returns, especially considering the improvements seen with STGS-T. Future work could also explore using alternative annealing schemes, or annealing with a different underlying estimator---\eg one could try a temperature-annealed GST.

From these results, considering both the maximum returns and the time to convergence, we note that alternative gradient estimation techniques can indeed yield better returns when incorporated into MADDPG, particularly the recently proposed GST, from~\citet{gappedStraightThrough}.

\subsection{Compute Time}
We now consider the computational requirements for each of the algorithms, using the toy problem outlined earlier. Recall that we perform these tests for three classes of estimator: STGS (which accounts for STGS\nobreakdash-1, STGS\nobreakdash-T, and TAGS); GRMC (with three different \(K\) values); and GST. Table~\ref{tab:compute-time-results} shows the outcome of these tests.

\newcommand{\FormatComputeResult}[3]{
    \begin{tabular}{@{}c@{}}
        \ensuremath{{#1}\pm{#2}}\\[-15pt]
        (\ensuremath{{#3}})
    \end{tabular}
}

\begin{table}[htb]
    \caption{Time-per-relaxation, in \(\mu\)s, for the three classes of gradient estimators, when using logits of various dimensionality as input. Results are given over a 95\% confidence interval from 5 different logit instances, where each procedure is repeated 10\,000 times. Underneath each metric, using round brackets, \((\cdot)\), we indicate how much slower the alternative techniques are, when compared to the baseline STGS.}
    \label{tab:compute-time-results}
    \centering
    \renewcommand{\arraystretch}{2.4}
    \setlength{\tabcolsep}{7pt}
    \resizebox{\linewidth}{!}{%
        \begin{filecontents*}{data/compute.csv}
theDIM,shouldwerule,theSTGS,theSTGSerr,theSTGSmult,theGRMCA,theGRMCAerr,theGRMCAmult,theGRMCB,theGRMCBerr,theGRMCBmult,theGRMCC,theGRMCCerr,theGRMCCmult,theGST,theGSTerr,theGSTmult
3,x,135.28,0.19,1.0,445.91,11.38,3.3,446.65,0.86,3.3,486.36,1.7,3.6,357.56,1.04,2.64
5,,135.65,0.51,1.0,438.52,0.63,3.23,446.66,0.7,3.29,501.9,1.16,3.7,356.95,0.6,2.63
10,,135.83,0.48,1.0,438.95,1.01,3.23,451.65,0.82,3.33,531.97,0.45,3.92,356.52,0.55,2.62
50,,134.05,0.89,1.0,440.46,1.01,3.29,484.77,1.29,3.62,765.56,2.09,5.71,356.77,1.77,2.66
100,,139.23,0.25,1.0,455.54,2.97,3.27,520.06,0.78,3.74,1055.18,4.16,7.58,359.26,1.26,2.58
1000,,154.17,0.45,1.0,533.12,1.01,3.46,1060.74,2.92,6.88,6171.59,8.67,40.03,386.57,1.36,2.51
\end{filecontents*}
\csvreader[
            head to column names,
            tabular=c||c|ccc|c,
            table head= \toprule
                \(\downarrow\) Dimensionality & \textbf{STGS} & \textbf{GRMC-1} & \textbf{GRMC-10} & \textbf{GRMC-50} & \textbf{GST},
            late after line=,
            before first line=,
            before line=\ifthenelse{\equal{\shouldwerule}{}}{\\}{\\\hline\hline},
            table foot = \\\bottomrule,
        ]{data/compute.csv}{}{%
            \theDIM &
            \FormatComputeResult{\theSTGS}{\theSTGSerr}{\theSTGSmult} &
            \FormatComputeResult{\theGRMCA}{\theGRMCAerr}{\theGRMCAmult} &
            \FormatComputeResult{\theGRMCB}{\theGRMCBerr}{\theGRMCBmult} &
            \FormatComputeResult{\theGRMCC}{\theGRMCCerr}{\theGRMCCmult} &
            \FormatComputeResult{\theGST}{\theGSTerr}{\theGSTmult}
        }
    }
\end{table}

We notice firstly that STGS scales well with dimensionality---the computational overhead when increasing the dimension does not change significantly. Even in the high-dimensional case of \(1\,000\), the technique is only marginally slower. These benefits are common to the baseline STGS\nobreakdash-1 approach, as well as the proposed techniques of STGS\nobreakdash-T and TAGS.

Next, we see that GRMC\(K\) is at least three times slower than the baseline approach. Moreover, using a larger \(K\) value does, understandably, increase the computational burden of the relaxation. Though this effect is not substantial for low-dimensional inputs, for higher-dimensional problems, \(K\) has a marked impact---\eg with \(K=50\), computation slows down considerably, becoming \(40\) times slower than the baseline for an input dimension of \(1\,000\).

The computational burden of GST sits somewhat in-between the baseline, STGS, and the GRMC\(K\) approach. Importantly, though, this method also scales well with dimensionality, staying at just over \(2.5\) times slower than the baseline, irrespective of the input size---an attractive property.

From these results, and the insights drawn from the previous section, we can draft general guidelines for choosing an alternative estimator: if minimising the computational burden is paramount for a given problem, it may be worth using the STGS\nobreakdash-T, for it has the same overhead as the STGS\nobreakdash-1, and it does yield improvements in both the achieved returns and convergence time. However, if one can afford a more expensive relaxation procedure, the GST is a good fit---it is somewhat slower, but the benefits are significant. Since GRMC\(K\) is more expensive than the GST, yet usually yields lower returns, it does not seem like a sensible option as an estimator in either case. Granted, it could be that better performance comes from increasing \(K\) further (\eg \citet{paulusRaoBlackwellizingStraightThroughGumbelSoftmax} use \(K=1000\) in some of their experiments), but the computational burden will only worsen in such a case, which is undesirable. TAGS, as it was presented here, should not be used.

\subsection{Gradient Variance}

\begin{figure}[htbp]
    \centering
    \includegraphics[width=\linewidth]{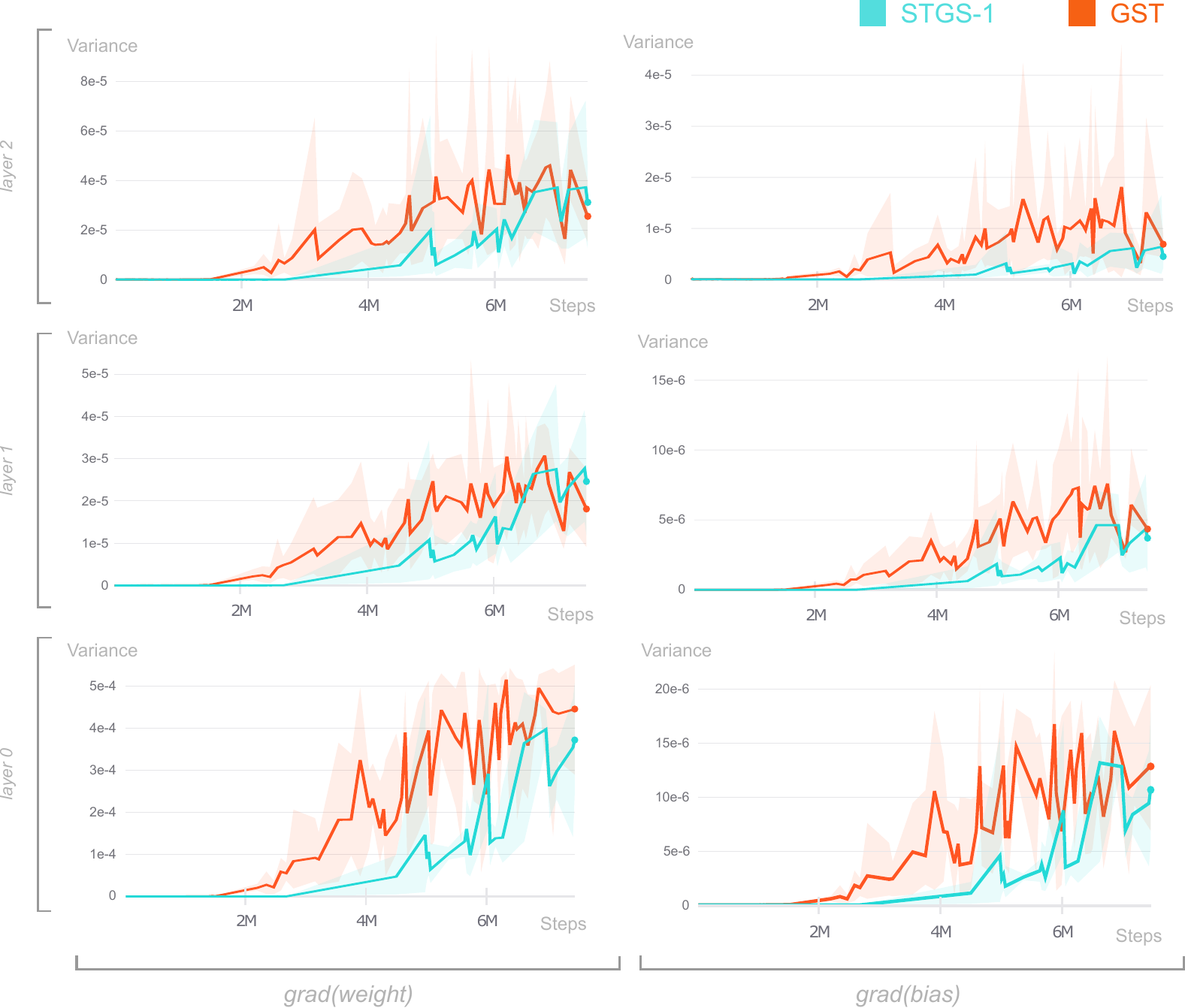}
    \caption{Plots showing the gradient variance (left: gradients of \emph{weight} parameters, right: gradients of \emph{bias} parameters), for each layer in the policy networks, for the 15x15-4p-5f task in LBF. The results are aggregated across the 4 agents in this task---the shaded region indicates the maximum and minimum values across agents, the solid line indicates the mean.}
    \label{fig:grad_variance}
\end{figure}

We have previously seen marked improvements in some tasks when using the proposed gradient estimators, particularly the GST. We now stimulate further discussion by presenting a cursory look into \emph{why}. We reconsider the LBF task of 15x15-4p-5f (see Figure~\ref{fig:returns:lbf-15x15-4p-5f}), and retrain with two algorithms: the baseline STGS-1, and the best performing alternative, GST. Figure~\ref{fig:grad_variance} shows the variance of the gradients across mini-batches, for each of the layers in the policy networks, over the course of the training.

Immediately, we notice a trend in these graphs: the variance of the gradients, taken across a mini-batch, increases more rapidly for the GST algorithm than those for the baseline. Though not definitive, such results indicate that more informative gradients are being propagated through the policy networks. Informative gradients, in turn, allow the algorithm to achieve higher returns and converge faster---as evidenced in the results seen previously.

\section{Related Work}

A core insight of this work is that discrete gradient estimation does not exist solely in the domain of RL. In fact, the original GS papers~\citep{jangCategoricalReparameterizationGumbelSoftmax,maddisonConcreteDistributionContinuous} demonstrated the technique on problems such as structured output prediction and density estimation. Indeed, discrete gradients arise in a wide variety of contexts---including discrete variational auto-encoders~\citep{rolfeDiscreteVariationalAutoencoders, kingma2019introduction}, hard attention~\citep{gulcehre2017memory, yan2018hierarchical}, generative adversarial networks for text~\citep{ganForDiscreteElements, pmlr-v70-zhang17b}, and convolutional networks~\citep{Veit2019ConvolutionalNW}. As a result, the biased reparameterisation of the GS is problematic in a wide variety of domains, and accordingly, a significant research effort has focused on improving the method. In this paper, we drew two methods from the literature: the GRMC\(K\)~\citep{paulusRaoBlackwellizingStraightThroughGumbelSoftmax} and the GST~\citep{gappedStraightThrough}. Various other approaches could, conceivably, be integrated into the MADDPG algorithm to handle discrete-action environments. We outline them briefly here.

The STGS, GRMC\(K\), and GST are all instances of the \emph{pathwise-derivative} approach to gradient estimation, which is underpinned by a more-general form of the Deterministic Policy Gradient Theorem, seen in (\ref{eqn:dpgt}). The Invertible Gaussian Reparameterisation~\citep{potapczynskiInvertibleGaussianReparameterization} also belongs to this class, where Gaussian noise is used instead of Gumbel noise. \citet{andriyashImprovedGradientBasedOptimization2019}, too, move away from using Gumbel noise in their method, and propose a simple piecewise-linear relaxation instead. Other methods steer further away from the STGS. For example, \citet{leeReparameterizationGradientNondifferentiable2018} generalise the reparameterisation trick through manifold sampling, and are able to create an unbiased and reduced-variance estimator. \citet{lorberbomDirectOptimizationArgmax2019} avoid the need to relax the categorical distribution altogether by applying the technique of \emph{direct optimisation}.

The other approach to gradient estimation is \emph{score-function} methods, which is underpinned by a more-general form of the Stochastic Policy Gradient Theorem, seen in (\ref{eqn:pgt}). Alternatives here include: subtracting a baseline~\citep{reinforceBaselinePaper, NIPS2001_584b98aa}; using a Taylor expansion of a mean-field network, as in MuProp~\citep{muprop}; and using copula-based sampling, as in CARMS~\citep{dimitrievCARMSCategoricalAntitheticREINFORCEMultiSample2021}. However, it is unclear how using a score-function method would change the underlying DPG mechanics when applied in MADDPG.

Authors have also combined score function and pathwise derivative methods, leveraging desirable qualities from both approaches. For example: using both REINFORCE and the GS in conjunction, as in REBAR~\citep{tuckerREBARLowvarianceUnbiased}; training a surrogate neural network as a control variate, as in RELAX~\citep{grathwohlBackpropagationVoidOptimizing}; and using sampling without replacement~\citep{koolEstimatingGradientsDiscrete}.

\section{Conclusion}
This paper explored the impact of the Gumbel-Softmax (GS) reparameterisation~\citep{jangCategoricalReparameterizationGumbelSoftmax,maddisonConcreteDistributionContinuous} on MADDPG~\citep{loweMultiAgentActorCriticMixed2017} when applied to grid-world environments. Firstly, some necessary theoretical foundations were presented and the problem was framed in the context of the broader literature. Thereafter, we looked closely at the straight-through GS (STGS) and discrete gradient estimation more generally, highlighting the key concepts therein. After presenting a handful of candidate STGS alternatives---two with simple tweaks to the STGS, and two from the literature~\citep{paulusRaoBlackwellizingStraightThroughGumbelSoftmax, gappedStraightThrough}---these estimators were implemented into the MADDPG algorithm. A suite of nine MARL tasks across two environments was used for testing, and various metrics were analysed.

On some of the tasks---particularly the simpler ones, where MADDPG already performed well---no significant changes were observed, in terms of returns achieved and the speed to convergence. On other tasks though, particularly in the more challenging ones, substantial improvements occurred. It was found that even an easy change to the original STGS estimator, simply lowering the temperature parameter, yielded good results. The proposed temperature-annealing scheme in TAGS, however, was shown to be a bad choice for the estimator---though we acknowledge a different set of hyperparameters may have helped here. The GRMC\(K\) estimator~\citep{paulusRaoBlackwellizingStraightThroughGumbelSoftmax} showed promising results, but was hindered by a below-par computational burden. Finally, far superior to the other methods was the GST estimator~\citep{gappedStraightThrough}. This method achieved the best results across a range of tasks, with up to 55\% higher returns, as well as faster convergence, when compared to the original STGS. Though it did introduce additional computational burden, at around 2.5 times slower than the STGS, the method nonetheless scaled well with dimensionality, and is certainly a viable technique for many use-cases.

We are now in a good position to support the suggestions made by \citet{papoudakisBenchmarkingMultiAgentDeep2021} in their benchmarking paper. Based on the empirical data observed, we agree that the bias of the STGS method is indeed problematic for MADDPG. As a result, by improving the estimator used---\ie by lowering its bias---we can improve the returns achieved by MADDPG. To answer our research question from Section~\ref{sect:intro}, then: yes, alternative discrete gradient-estimation techniques \emph{can} improve the performance of MADDPG in discrete grid-worlds.

Notice the benefit of our findings. We can take the extant MADDPG algorithm, replace \emph{only} the gradient estimation technique---that is, swap out, \eg, the STGS for the GST, and leave everything else the same---and the resulting performance may likely improve. Though our algorithm becomes slightly more expensive computationally, we witness faster convergence and higher returns, with minimal development overhead.

\section{Future Work}

Many avenues of future work extend from this paper---we highlight a handful here.

Firstly, it would be useful to evaluate the performance of the proposed algorithms on a wider variety of tasks---in particular, the results in RWARE were promising but distinctly limited, with only two tasks tested. Moreover, only co-operative tasks were tested here, and adversarial configurations should be explored too.

Secondly, much more investigation ought to be done into why the GST is boasting better performance. Though an interesting foray, the analysis into the gradient variance was just a first step. Future research should continue to focus on the mechanics of the algorithms, and probe at various points.

Thirdly, the core MADDPG algorithm designed for this paper did not incorporate various extensions suggested in the literature, \eg parameter sharing~\citep{christianos2021scaling, christianosSharedExperienceActorCritic}. Combining the benefits observed here with other strong extensions elsewhere would be an interesting exercise. Furthermore, a wider hyperparameter search, now with the alternative estimators involved too, may be helpful.

Finally, we note that only two alternative methods from the literature were presented here---the GRMC\(K\)~\citep{paulusRaoBlackwellizingStraightThroughGumbelSoftmax} and the GST~\citep{gappedStraightThrough}. Though sufficient for our analysis, it would be useful to explore the other options synthesised from the literature (\eg~\citep{andriyashImprovedGradientBasedOptimization2019, leeReparameterizationGradientNondifferentiable2018, lorberbomDirectPolicyGradients2020}). Some of these, though more complex, boast many attractive properties, and may prove to be even more fruitful.

\bibliographystyle{ACM-Reference-Format} 
\bibliography{references}

\end{document}